# Constrained Bayesian Inference for Low Rank Multitask Learning

**Oluwasanmi Koyejo**
Electrical Engineering Dept.
University of Texas at Austin

**Joydeep Ghosh**
Electrical Engineering Dept.
University of Texas at Austin

## Abstract

We present a novel approach for constrained Bayesian inference. Unlike current methods, our approach does not require convexity of the constraint set. We reduce the constrained variational inference to a parametric optimization over the feasible set of densities and propose a general recipe for such problems. We apply the proposed constrained Bayesian inference approach to multitask learning subject to rank constraints on the weight matrix. Further, constrained parameter estimation is applied to recover the sparse conditional independence structure encoded by prior precision matrices. Our approach is motivated by reverse inference for high dimensional functional neuroimaging, a domain where the high dimensionality and small number of examples requires the use of constraints to ensure meaningful and effective models. For this application, we propose a model that jointly learns a weight matrix and the prior inverse covariance structure between different tasks. We present experimental validation showing that the proposed approach outperforms strong baseline models in terms of predictive performance and structure recovery.

## 1 INTRODUCTION

The Bayesian paradigm has become one of the most important approaches for modeling uncertainty. Bayes' classic theorem provides a principle for updating prior beliefs with new information, and has become and important component of statistics and machine learning. Despite the elegance of Bayes theorem, some learning problems require model constraints that are difficult or inappropriate to enforce using standard prior distributions. Examples include linear inequality constraints (Gelfand et al., 1992) and margin constraints (Zhu et al., 2012).

Williams (1980) showed that the Bayesian posterior distribution can be derived as the solution of a constrained relative entropy minimization problem. Constrained Bayesian inference is proposed as an extension that combines Bayesian inference with other constraints determined by domain knowledge. Constrained relative entropy minimization can be solved via Fenchel duality theory (Altun & Smola, 2006), which requires convexity of the constraint set. In this paper, we propose an alternative approach that does not require the constraint set to be convex. We find that the optimization problem resulting from the proposed approach may be easier to solve than the equivalent Fenchel dual approach even when the constraint set is convex.

Rank constraints have proven to be effective for controlling model complexity (Candés & Recht, 2009). In addition to superior performance in many scenarios, the decompositions learned are often useful for explaining the structure of complex multivariate data. Low rank latent variable models have been applied to various domains including principal component analysis (Bishop, 1998), multitask learning (Stegle et al., 2011) and collaborative filtering (Salakhutdinov & Mnih, 2008). We propose a novel approach for low rank multitask learning via Bayesian inference subject to a nuclear norm constraint on the predictive weight matrix. The constrained inference is combined with parameter estimation for the prior precision of the matrix-variate Gaussian distribution. We enforce $l_1$ regularization constraints on the precision matrix to reveal its sparsity structure.

Our work is motivated by reverse inference for functional neuroimaging. Functional neuroimaging datasets typically consist of a relatively small number of correlated high dimensional brain images. Hence, capturing the inherent structural properties of the imaging data is critical for robust inference. Predictive modeling (also known as "brain reading" or "reverse inference") has become an increasingly popular approach for studying fMRI data (Pereira et al., 2009; Poldrack, 2011). Reverse inference involves the interpretation of the parameters of a model trained to decode the stimulus or task using the brain images as features. We show that the proposed approach is effective in this domain.

The contributions of this paper are as follows:

- We propose an a novel representation approach for constrained Bayesian inference. We prove that the optimizing density is a member of an exponential family. The presented results relax the necessary conditions on the constraint set such as convexity.

- We develop a novel constrained Bayesian model for rank constrained multitask learning and apply $l_1$ norm constrained parameter estimation to estimate the intertask conditional independence structure.

- The proposed multitask learning approach is applied to reverse inference for functional neuroimaging data. We show that the proposed approach results in superior accuracy as compared to strong baseline models.

As a minor contribution, our work appears to be the first application of constrained Bayesian inference to continuous valued variables that are not margin constraints. Constrained Bayesian inference is discussed in Section 2 and the proposed representation approach is introduced in Section 2.1. We discuss the proposed rank constrained multitask learning approach in Section 3. Related work is discussed in Section 4 and experimental results are presented in Section 5.

## 1.1 PRELIMINARIES

We denote vectors by lower case $\mathbf{x}$ and matrices by capital $\mathbf{X}$. Let $\mathbf{I}_D$ represent the $D \times D$ identity matrix. Given a matrix $\mathbf{A} \in R^{P \times Q}$, $\text{vec}(\mathbf{A}) \in R^{PQ}$ is the vector obtained by concatenating columns of $\mathbf{A}$. Given matrices $\mathbf{A} \in R^{P \times Q}$ and $\mathbf{B} \in R^{P' \times Q'}$, the Kronecker product of $\mathbf{A}$ and $\mathbf{B}$ is denoted as $\mathbf{A} \otimes \mathbf{B} \in R^{PP' \times QQ'}$. We use $\|\cdot\|_p$ to denote the vector $L_p$ norm with $\|\mathbf{x}\|_p = (\sum_i x_i^p)^{\frac{1}{p}}$, and use $\|\|\cdot\|\|$ to denote spectral (matrix) norms i.e. $\|\|\mathbf{X}\|\|_p$ is the $L_p$ norm of the singular values of $\mathbf{X}$.

Let $\mathcal{X}$ be a Banach space and let $\mathcal{X}^*$ be the *dual space* of $\mathcal{X}$. The *Legendre-Fenchel* transformation (or convex conjugate) of a function $f : \mathcal{X} \mapsto [-\infty, +\infty]$ is given by $f^* : \mathcal{X}^* \mapsto [-\infty, +\infty]$ as $f^*(x^*) = \sup_{x \in \mathcal{X}} \{\langle x, x^* \rangle - f(x)\}$.
where $\langle x, x^* \rangle$ denotes the dual pairing. See Borwein & Zhu (2005) for further details on Fenchel duality, particularly as applied to variational optimization.

Let $E$ be the *expectation operator* with $\mathrm{E}_p[f(z)] = \int_z p(z)f(z)dz$. The *Kullback-Leibler divergence* between densities $q$ and $p$ is given by $\mathrm{KL}(q(z)\|p(z)) = \mathrm{E}_{q(z)}[\log q(z) - \log p(z)]$. The *delta function* as a generalized function that satisfies $\int_Z f(z)\delta_a\{dz\} = f(a)$, where $f$ is absolutely continuous with respect to $dz$, and $a \in Z$. Following from the definition, the expectation with respect to the delta function satisfies $\mathrm{E}_{\delta_a}[f] = f(a)$, and given the density $p$, we have that $\mathrm{E}_p[\delta_a] = p(a)$. Further, it can be shown (Williams, 1980) that $\mathrm{KL}(\delta_a\|p) = -\log p(a)$.

An *exponential family* is a class of probability distributions whose density functions take the form:

$$p(\mathbf{x}|\boldsymbol{\theta}) = h(\mathbf{x})e^{\langle \boldsymbol{\eta}(\boldsymbol{\theta}), \mathbf{t}(\mathbf{x}) \rangle - G(\boldsymbol{\theta})},$$

where $\boldsymbol{\eta}(\boldsymbol{\theta})$ is known as the natural parameter vector, $\mathbf{t}(\mathbf{x})$ is the vector of natural statistics, $G(\boldsymbol{\theta})$ is the log partition function and $h(\mathbf{x})$ is known as the base measure. The exponential family is in *canonical* form if $\boldsymbol{\eta}(\boldsymbol{\theta}) = \boldsymbol{\theta}$. Further details on exponential family distributions may be found in (Brown, 1986).

Let $\mathbf{x} \in R^D$ be drawn from a *multivariate Gaussian* distribution. The density is given as:

$$\mathcal{N}(\mathbf{m}, \boldsymbol{\Sigma}) = \frac{\exp\left(-\frac{1}{2}\mathrm{tr}\left((\mathbf{x} - \mathbf{m})^\top \boldsymbol{\Sigma}^{-1} (\mathbf{x} - \mathbf{m})\right)\right)}{(2\pi)^{D/2}|\boldsymbol{\Sigma}|^{P/2}},$$

where $\mathbf{m} \in R^D$ is the mean vector and $\boldsymbol{\Sigma} \in R^{D \times D}$ is the covariance matrix. $|\cdot|$ denotes the matrix determinant and $\mathrm{tr}(\cdot)$ denotes the matrix trace. Let $\mathbf{X} \in R^{D \times K}$ be drawn from a *matrix-variate Gaussian* distribution represented as $\mathcal{MN}(\mathbf{M}, \boldsymbol{\Sigma}_R, \boldsymbol{\Sigma}_C)$ where $\mathbf{M} \in R^{D \times K}$ is the mean matrix, $\boldsymbol{\Sigma}_R \in R^{D \times D}$ is the row covariance matrix and $\boldsymbol{\Sigma}_C \in R^{K \times K}$ is the column covariance matrix. The density is given by:

$$\frac{\exp\left(-\frac{1}{2}\mathrm{tr}\left(\boldsymbol{\Sigma}_C^{-1}(\mathbf{X} - \mathbf{M})^\top \boldsymbol{\Sigma}_R^{-1}(\mathbf{X} - \mathbf{M})\right)\right)}{(2\pi)^{DL/2}|\boldsymbol{\Sigma}_R|^{D/2}|\boldsymbol{\Sigma}_C|^{K/2}}.$$

## 2 CONSTRAINED BAYESIAN INFERENCE

Constrained relative entropy inference follows from the *principle of minimum discrimination information* (Kullback, 1959); a conceptual framework for updating a distribution given constraints. It defines a procedure for updating the distribution as one that satisfies the constraints and is closest to a predefined prior distribution in terms of relative entropy. Bayesian inference is recovered from the constrained relative entropy framework when the data constraints correspond to knowledge of the value of $y$ with certainty (Williams, 1980). Given the observation $\tilde{y} \sim P_y$, this knowledge is encoded using the constraint $\mathrm{E}_q[\delta_{\tilde{y}}] = 1$ that must be satisfied by the updated distribution $q$. The resulting constrained relative entropy minimization problem is given by:

$$\min_{q \in \mathcal{P}} \left[ \mathrm{KL}(q(z,y)\|p(z,y)) \text{ s.t. } \mathrm{E}_q[\delta_{\tilde{y}}] = 1 \right]. \quad (1)$$

It is clear that any distribution $q$ that optimizes (1) must satisfy the equivalent conditional distribution constraint

$q(y) = \int_Z q(z,y) = \delta_{\tilde{y}}$, so we will focus on estimating the portion of the distribution that remains unknown, which, from the basic rules of probability, is the conditional distribution $q(z|y)$. Thus, it will be useful to express the joint relative entropy in a form that separates the latent variables from the observations by expressing $\mathrm{KL}(q(z,y)\|p(z,y))$ as:

$$\mathrm{E}_{q(y)}\left[\mathrm{KL}(q(z|y)\|p(z|y))\right] + \mathrm{KL}(q(y)\|p(y)).$$

Enforcing the constraint $q(y) = \delta_{\tilde{y}}$, we recover that $\mathrm{KL}(q(x|y)q(y)\|p(x|y)p(y))$ is given by:

$$\mathrm{KL}(q(x|y=\tilde{y})\|p(x|y=\tilde{y})) - \log(p(\tilde{y})).$$

The second term $\log(p(\tilde{y}))$ is the log evidence, and is fixed independent of the first term. The first term is minimized when $q(x|y=\tilde{y}) = p(x|y=\tilde{y})$, recovering the Bayesian posterior distribution[1]. Thus, the solution of the relative entropy minimization problem (1) takes the form of the generalized density $q_*(x,y) = p(x|y=\tilde{y})\delta_{\tilde{y}}$.

For the rest of the discussion, we focus on $q$ as a density with respect to $Z$ (ignoring the implicit conditioning). It is instructive to expand the terms of the loss function. The relative entropy expands as follows:

$$\mathrm{KL}(q(z)\|p(z|y)) - \log p(y) \tag{2a}$$
$$= \mathrm{E}_q\left[\log q(z) - \log p(z|y) - \log p(y)\right] \tag{2b}$$
$$= \mathrm{E}_q\left[\log q(z) - \log p(z) - \log p(y|z)\right] \tag{2c}$$
$$= \mathrm{KL}(q(z)\|p(z)) - \mathrm{E}_q\left[\log p(y|z)\right]. \tag{2d}$$

where (2b) and (2d) follow directly by expansion of the KL divergence, and (2c) follows from the rules of conditional probability as $p(z|y)p(y) = p(z,y) = p(y|z)p(z)$. The result of (2c) also recovers the identity discovered by Zellner (1988), who showed that the Bayesian posterior density is given by:

$$p(z|y) = \arg\min_{q \in \mathcal{P}} \mathrm{KL}(q(z)\|p(z)) - \mathrm{E}_q\left[\log p(y|z)\right]. \tag{3}$$

Constrained Bayesian inference defines a procedure for enforcing constraints on latent variables in addition to the constraints on the observation variables. Let $\boldsymbol{\beta}$ represent *feature functions* that map $Z$ to a feature space with components $\boldsymbol{\beta}(z) = \{\beta_j(z)\}$ and let $\mathsf{C}$ denote a constraint set of interest. We consider information encoded as expectation constraints in this paper. The *constrained Bayesian inference* procedure is defined by the following equivalent optimization problems:

$$\min_{q \in \mathcal{P},\, \mathrm{E}_q[\boldsymbol{\beta}(z)] \in \mathsf{C}} \left[\mathrm{KL}(q(z)\|p(z|y))\right] \tag{4a}$$

$$\min_{q \in \mathcal{P},\, \mathrm{E}_q[\boldsymbol{\beta}(z)] \in \mathsf{C}} \left[\mathrm{KL}(q(z)\|p(z)) - \mathrm{E}_q\left[\log p(y|z)\right]\right] \tag{4b}$$

---
[1] Recall Bayes rule: $p(z|y) = p(y|z)p(z)/p(y)$

It is clear from (4a) that constrained Bayesian inference corresponds to an information projection of the Bayesian posterior distribution to the set to distributions $q$ that satisfy the constraints $\mathrm{E}_q\left[\boldsymbol{\beta}(z)\right] \in \mathsf{C}$. Following Zellner, we call $q_*$ the *postdata* distribution to distinguish it from the unconstrained Bayesian posterior distribution.

We now consider probabilistic inference via constrained relative entropy minimization. Altun & Smola (2006) studied norm ball constraints given by $\|\mathrm{E}_q\left[\boldsymbol{\beta}(z)\right] - \mathbf{b}\|_\mathcal{B} \leq \epsilon$ where $\|\cdot\|_\mathcal{B}$ is the norm ball on a the Banach space $\mathcal{B}$ centered at $\mathbf{b} \in \mathcal{B}$, and $\epsilon \geq 0$ is the width. The solution was found by an elegant application of Fenchel duality for variational optimization (Borwein & Zhu, 2005). The following Lemma characterizes relative entropy minimization subject to norm ball constraints.

**Lemma 1** (Altun & Smola (2006)).

$$\min_{q \in \mathcal{P}} \mathrm{KL}(q(z)\|p(z)) \text{ s.t. } \|\mathrm{E}_q\left[\boldsymbol{\beta}(z)\right] - \mathbf{b}\|_\mathcal{B} \leq \epsilon \tag{5}$$

$$= \max_{\boldsymbol{\lambda}} \langle \boldsymbol{\lambda}, \mathbf{b} \rangle - \log \int_Z p(z)e^{\langle \boldsymbol{\lambda}, \boldsymbol{\beta}(z) \rangle} dz - \epsilon\|\boldsymbol{\lambda}\|_{\mathcal{B}^*} + e^{-1} \tag{6}$$

*and the unique solution is given by* $q_*(z) = p(z)e^{\langle \boldsymbol{\lambda}_*, \boldsymbol{\beta}(z)\rangle - G(\boldsymbol{\lambda}_*)}$ *where* $\boldsymbol{\lambda}_*$ *is the solution of the dual optimization* (6) *and* $G(\boldsymbol{\lambda}_*)$ *ensures normalization.*

There may be several equivalent representations for a given density $q \in \mathcal{P}$. However, Lemma 1 shows that the density that minimizes relative entropy subject to norm ball constraints, if it exists, has a canonical representation a member of the exponential family with base measure $p$, natural statistics $\boldsymbol{\beta}(z)$ and parameters $\boldsymbol{\lambda}_*$. The conditions for Lemma 1 include constraint qualification, which requires the existence of densities that satisfy the set of constraints, and a finite cost (6) at the solution $\boldsymbol{\lambda}_*$. More details are given in Altun & Smola (2006) and Chapter 4 of Borwein & Zhu (2005).

### 2.1 A REPRESENTATION APPROACH

The dual solution presented in Lemma 1 requires convexity of the constraint set $\mathsf{C}$. Further, solving the resulting dual optimization (6) requires the evaluation of the log partition function which is often challenging. We present an alternative representation approach that separates the problem into two parts. First we find the parametric family of the optimizing postdata density, then we directly optimize over that parametric family. Unlike the dual approach, the proposed representation approach does not require convexity of the constraint set.

For the rest of this paper, we will assume that the set of solutions $q_*$ of the constrained Bayesian optimization (4) is not empty so the optimization problem is well defined. This implies the existence of at least one density $q \in \mathcal{P}$ that satisfies the constraints $\mathrm{E}_q\left[\boldsymbol{\beta}(z)\right] \in \mathsf{C}$. This also implies that the solution of the variational optimization problem is achieved at a density $q_*$. Further, we assume that for each solution $q_*$, the expectation $\mathrm{E}_{q_*}\left[\boldsymbol{\beta}(z)\right] = \mathbf{a}_*$ is bounded to avoid the degenerate problem of unbounded constraints. Finally, we assume that $\mathsf{C} \subset \mathcal{B}$ is a closed subset of the Banach space $\mathcal{B}$. This assumption is mostly for convenience and clarity and can easily be relaxed.

Let $\mathcal{E}_{\mathbf{c}} = \{q \in \mathcal{P} \,|\, \mathrm{E}_q\left[\boldsymbol{\beta}(z)\right] = \mathbf{c}\}$ denote the constraint set subject to equality constraints. The constrained Bayes optimization problem (4b) can be written as:

$$\min_{\mathbf{c} \in \mathsf{C}} \left[\min_{q \in \mathcal{E}_{\mathbf{c}}} \mathrm{KL}(q(z) \| p(z|y))\right], \tag{7}$$

which requires the solution of an inner optimization:

$$q_{\mathbf{c}} = \arg\min_{q \in \mathcal{E}_{\mathbf{c}}} \mathrm{KL}(q(z) \| p(z|y)). \tag{8}$$

Let $\mathsf{A} \subset \mathsf{C}$ represent the set of points $\mathbf{c} \in \mathsf{C}$ where $\mathbf{c}$ is bounded, and the optimization problem of (4b) is finite and attained. Assuming the existence of at least one solution $q_*$, it follows that the set $\mathsf{A}$ is not empty. We associate a density function $q_{\mathbf{c}}$ to every element $\mathbf{c} \in \mathsf{A}$. We define a *feasible set* of solutions characterized by the set of densities $\mathcal{F} = \{q_{\mathbf{c}}(z) \,|\, \mathbf{c} \in \mathsf{A}\}$. The following proposition is a direct consequence of Lemma 1 and is stated without proof.

**Proposition 2.** *For any $\mathbf{c} \in \mathsf{A}$, the unique minimizer of (8) is given by:* $q_{\mathbf{c}}(z) = p(z|y) e^{\langle \boldsymbol{\lambda}_{\mathbf{c}}, \boldsymbol{\beta}(z) \rangle - G(\boldsymbol{\lambda}_{\mathbf{c}})}$ *where $\boldsymbol{\lambda}_{\mathbf{c}}$ is the solution of the dual optimization (6) with $\epsilon = 0$ and $G(\boldsymbol{\lambda}_{\mathbf{c}})$ ensures normalization.*

We may now state the main result.

**Theorem 3.** *Let $\mathcal{F} = \{q_{\mathbf{c}} \,|\, \mathbf{c} \in \mathsf{A}\}$ denote the feasible set of (8). The postdata density given by the minimizer of (4a) is the solution of:*

$$q_* = \arg\min_{q \in \mathcal{F}} \mathrm{KL}(q(z) \| p(z|y))$$

*and the solution is given by $q_* = q_{\mathbf{a}_*}$ for the optimal $\mathbf{a}_* \in \mathsf{A}$ with $q_*(z) = p(z|y) e^{\langle \boldsymbol{\lambda}_{\mathbf{a}_*}, \boldsymbol{\beta}(z) \rangle - G(\boldsymbol{\lambda}_{\mathbf{a}_*})}$ where $\boldsymbol{\lambda}_{\mathbf{a}_*}$ is the solution of the dual optimization (6) with the constraint set $\mathsf{C}' = \{\mathrm{E}_q\left[\boldsymbol{\beta}(z)\right] = \mathbf{a}_*\}$ and $G(\boldsymbol{\lambda}_{\mathbf{a}_*})$ ensures normalization. The solution is unique if $\mathsf{A}$ is convex.*

*Sketch of proof:* First we prove that $q_* \in \mathcal{F}$ by contradiction. Suppose $q_* \notin \mathcal{F}$, then $q_* = q_{\mathbf{v}}$ for $\mathbf{v} \notin \mathsf{A}$. This is a contradiction by definition of $\mathsf{A}$. The first claim follows directly. The parametric form of the solution follows from Proposition 2 and the uniqueness of the solution for convex $\mathsf{A}$ follows from the strict convexity of the relative entropy.

Applied directly, Theorem 3 requires the solution of the equality constrained variational inference (8) in the inner loop. The key insight from Proposition 2 is that the solution of (8) fully specifies the parametric form of the density. In other words, all the members of the set $\mathcal{F} = \{q_{\mathbf{c}} \,|\, \mathbf{c} \in \mathsf{A}\}$ have the same parametric form $f$ where $q_{\mathbf{c}} = f_{\boldsymbol{\theta}_{\mathbf{c}}}(z)$ is determined by the choice of $\mathbf{c}$. By definition of the exponential family, all $\boldsymbol{\theta} \in \boldsymbol{\Theta}$ where $\boldsymbol{\Theta}$ is the constraint set of the parametric distribution family containing $f_{\boldsymbol{\theta}}$.

**Corollary 4.** *The postdata density is the minimizer of (4a) and is given by $q_* = f_{\boldsymbol{\theta}_*}$ where $\boldsymbol{\theta}_*$ is the solution of:*

$$\boldsymbol{\theta}_* = \arg\min_{\boldsymbol{\theta} \in \boldsymbol{\Theta}} \left[ \begin{array}{c} \mathrm{KL}(f_{\boldsymbol{\theta}}(z) \| p(z|y)) \\ \text{s.t. } \mathrm{E}_{f_{\boldsymbol{\theta}}}\left[\boldsymbol{\beta}(z)\right] \in \mathsf{C} \end{array} \right].$$

*Sketch of proof:* Let the exponential family $\mathcal{G} = \{f_{\boldsymbol{\theta}} \,|\, \forall \boldsymbol{\theta} \in \boldsymbol{\Theta}\}$. Clearly $\mathcal{F} \subseteq \mathcal{G}$ by definition. Suppose $f_{\boldsymbol{\theta}_*} \notin \mathcal{F}$, feasibility implies that $f_{\boldsymbol{\theta}_*}$ satisfies the constraints. Thus $\exists \mathbf{v}$ such that $\mathrm{E}_{f_{\boldsymbol{\theta}_*}}\left[\boldsymbol{\beta}(z)\right] = \mathbf{v}$ and $\mathbf{v} \notin \mathsf{A}$. This is a contradiction by definition of $\mathsf{A}$. Thus $f_{\boldsymbol{\theta}_*} = q_* \in \mathcal{F} \subset \mathcal{G}$ and the proof follows from Theorem 3.

Our approach suggests the following *recipe* for constrained Bayesian inference. First, Proposition 2 is applied to specify the parametric form of $q_*$, then Corollary 4 is applied to convert the variational problem into a parametric optimization problem.

## 3 RANK CONSTRAINED MULTITASK LEARNING

Let $n = 1 \ldots N$ denote the number of training examples and $k = 1 \ldots K$ denote each task so that the output is given by $y_{n,k} \in R$. Given a $D$ dimensional feature vector $\mathbf{x}_n \in R^D$ and a weight vector $\mathbf{w}_k \in R^D$, each output is generated as:

$$y_{n,k} = \mathbf{w}_k^\top \mathbf{x}_n + \epsilon$$

where $\epsilon \sim \mathcal{N}\left(0, \sigma^2\right)$. The outputs may be collected into a output matrix $\mathbf{Y} \in R^{N \times K}$ and the features may be collected into a feature matrix $\mathbf{X} \in R^{D \times K}$ with $\mathbf{X}(n) = \mathbf{x}_n^\top$. The latent matrix $\mathbf{W}$ is drawn from a zero mean matrix-variate Gaussian distribution $\mathbf{W} \sim \mathcal{MN}\left(\mathbf{0}, \mathbf{R}, \mathbf{C}\right)$ with row covariance $\mathbf{R} \in R^{D \times D}$ and column covariance matrix $\mathbf{C} \in R^{K \times K}$. Fig. 1 illustrates the combined generative model. Without loss of generality, we assume that the output matrix is normalized to zero mean over the columns so we do not include a bias term. The model parameters are given by $\boldsymbol{\Theta} = \{\mathbf{R}, \mathbf{C}, \sigma^2\}$.

The unconstrained posterior distribution can be computed in closed form (Bishop, 2006). Let $\mathbf{w} = \mathrm{vec}(\mathbf{W})$ and $\mathbf{y} = \mathrm{vec}(\mathbf{Y})$, then $p(\mathbf{w}|\mathbf{y}) = \mathcal{N}(\boldsymbol{\mu}, \boldsymbol{\Sigma})$ where:

$$\boldsymbol{\mu} = \frac{1}{\sigma^2} \boldsymbol{\Sigma}(\mathbf{I}_K \otimes \mathbf{X}^\top)\mathbf{y} \tag{9}$$

$$\boldsymbol{\Sigma}^{-1} = (\mathbf{C}^{-1} \otimes \mathbf{R}^{-1}) + \frac{1}{\sigma^2}(\mathbf{I}_K \otimes \mathbf{X}^\top \mathbf{X}) \tag{10}$$

with $\boldsymbol{\mu} \in R^{DK}$ and $\boldsymbol{\Sigma} \in R^{DK \times DK}$.

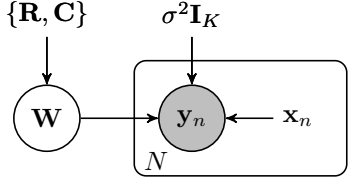

Figure 1: Generative Model for Multitask Learning

### 3.1 CONSTRAINED INFERENCE

We seek to enforce a rank constraint via the constraint set $\mathsf{B} = \{\mathbf{B} \,|\, \text{rank}(\mathbf{B}) \leq R\}$. We apply the recipe discussed in Section 2.1. First we must define the *parametric form* of the postdata distribution by solving (8) for a fixed $\mathbf{b} \in \mathsf{B}$. We find that $q_\mathbf{b}(\mathbf{w})$ is Gaussian distributed with density $\mathcal{N}(\mathbf{m}, \mathbf{S})$ and $\mathbf{m} = \mathbf{b}$. Following the arguments of Corollary 4, the postdata distribution is found by minimizing the KL divergence between the Gaussian distribution $\mathcal{N}(\mathbf{m}, \mathbf{S})$ and the Bayesian posterior distribution $\mathcal{N}(\boldsymbol{\mu}, \boldsymbol{\Sigma})$. This is given by:

$$\min_{\mathbf{m} \in \mathsf{B}, \mathbf{S}} \text{tr}(\boldsymbol{\Sigma}^{-1}\mathbf{S}) + (\boldsymbol{\mu} - \mathbf{m})^\top \boldsymbol{\Sigma}^{-1}(\boldsymbol{\mu} - \mathbf{m}) \\ - \log|\mathbf{S}| + \log|\boldsymbol{\Sigma}| \quad (11)$$

where $\mathbf{m} = \text{vec}(\mathbf{M})$. The optimization decouples between the mean term $\mathbf{m}$ and the covariance term $\mathbf{S}$. The minimum in terms of the covariance is achieved for $\mathbf{S} = \boldsymbol{\Sigma}$ and the mean optimization is given by the solution of a rank constrained quadratic optimization. We note that the Gaussian form of the constrained postdata density was not assumed *a-priori*, but was found as the solution to the constrained inference.

The solution of (11) requires computation and storage of the posterior covariance $\boldsymbol{\Sigma}$. This may become computationally infeasible for high dimensional data. In such situations, it may be more computationally efficient to estimate the postdata mean matrix using the form of (4b). Ignoring terms independent of the mean, this results in the optimization problem:

$$\min_{\mathbf{M} \in \mathsf{B}} \frac{1}{\sigma^2} \|\mathbf{Y} - \mathbf{XM}\|_2^2 + \text{tr}(\mathbf{M}^\top \mathbf{R}^{-1} \mathbf{M} \mathbf{C}^{-1}) \quad (12)$$

**Nuclear norm constraint:** The rank constraint is non-convex and is challenging to optimize directly. To simplify the optimization, we replace the rank constraint with a nuclear norm constraint $\mathsf{D} = \{\mathbf{D} \,|\, \|\mathbf{D}\|_1 \leq C\}$. The nuclear norm is computed as sum of the singular values of the matrix i.e. $\|\mathbf{D}\|_1 = \sum \sigma_i(\mathbf{D})$ where $\sigma_i(\mathbf{D})$ is the $i^{th}$ singular value of the matrix $\mathbf{D}$. The nuclear norm is known to encourage low rank solutions Candés & Recht (2009). The resulting postdata mean inference retains the same form with the new constraint set. Replacing the constraint set with a regularization function, we find that the postdata mean optimization can be rewritten as:

$$\min_{\mathbf{M}} \frac{1}{\sigma^2} \|\mathbf{Y} - \mathbf{XM}\|_2^2 + \text{tr}(\mathbf{M}^\top \mathbf{R}^{-1} \mathbf{M} \mathbf{C}^{-1}) + \|\mathbf{M}\|_1 \quad (13)$$

We note that there is no need for a regularization parameter if we learn the hyperparameters $\boldsymbol{\Theta} = \{\mathbf{R}, \mathbf{C}, \sigma^2\}$, as the optimization only depends on the relative scale of the three terms.

**Kronecker Covariance constraint:** Unlike the prior covariance, the posterior covariance matrix does not decompose into Kronecker form. Hence, the size of the posterior covariance may be of computational concern. We propose a Kronecker factorization constraint structure for the posterior covariance matrix. Following Theorem 3, we find that the postdata distribution retains its Gaussian form. Let $\mathbf{S} = \mathbf{H} \otimes \mathbf{G}$ where $\mathbf{G} \in R^{D \times D}$ is constrained row covariance matrix and $\mathbf{H} \in R^{K \times K}$ is the constrained column covariance matrix. Employing the cost function (4b) and ignoring terms independent of the postdata covariance, we compute:

$$\min_{\mathbf{G}, \mathbf{H}} \frac{1}{\sigma^2} \text{tr}(\mathbf{X}^\top \mathbf{X} \mathbf{G}) \text{tr}(\mathbf{H}) + \text{tr}(\mathbf{R}^{-1}\mathbf{G}) \text{tr}(\mathbf{C}^{-1}\mathbf{H}) \\ - K \log|\mathbf{G}| - D \log|\mathbf{H}|$$

This can be solved using an alternating optimization approach:

$$\mathbf{G}^{-1} = \frac{1}{K} \left( \frac{\text{tr}(\mathbf{H})}{\sigma^2} \mathbf{X}^\top \mathbf{X} + \text{tr}(\mathbf{C}^{-1}\mathbf{H}) \mathbf{R}^{-1} \right) \quad (14)$$

$$\mathbf{H}^{-1} = \frac{1}{D} \left( \frac{\text{tr}(\mathbf{X}^\top \mathbf{X} \mathbf{G})}{\sigma^2} \mathbf{I}_K + \text{tr}(\mathbf{R}^{-1}\mathbf{G}) \mathbf{C}^{-1} \right) \quad (15)$$

The result of constrained inference is the postdata distribution $q_*(\mathbf{W}|\mathbf{Y}) = \mathcal{MN}(\mathbf{M}, \mathbf{G}, \mathbf{H})$.

### 3.2 PARAMETER ESTIMATION

In addition to low rank constraints on the weight matrix, we are interested in learning the prior conditional independence structure between the features and between the tasks. This is achieved by placing Laplacian priors (Friedman et al., 2008; Stegle et al., 2011) on the row and column prior precision matrices:

$$p(\mathbf{R}^{-1}) \propto \exp(-\lambda_r \|\mathbf{R}^{-1}\|_1)[\mathbf{R}^{-1} \succ 0],$$
$$p(\mathbf{C}^{-1}) \propto \exp(-\lambda_c \|\mathbf{C}^{-1}\|_1)[\mathbf{C}^{-1} \succ 0],$$

where the $l_1$ norm is given by $\|\mathbf{R}\|_1 = \sum_{i,j} |r_{ij}|$. Ignoring terms independent of the precision matrices, the loss function is given by:

$$\min_{\mathbf{R}^{-1}, \mathbf{C}^{-1}} \text{tr}(\mathbf{R}^{-1}\mathbf{G}) \text{tr}(\mathbf{C}^{-1}\mathbf{H}) + \text{tr}(\mathbf{W}^\top \mathbf{R}^{-1} \mathbf{W} \mathbf{C}^{-1}) \\ - K \log|\mathbf{R}| - D \log|\mathbf{C}| + \lambda_r \|\mathbf{R}^{-1}\|_1 + \lambda_c \|\mathbf{C}^{-1}\|_1 \quad (16)$$

**Algorithm 1** Constrained Inference and Parameter Estimation for Multitask Learning

    **Initialize** $G, H, \Theta = \{R, C, \sigma^2\}$
    **repeat**
        Update $M|\Theta$ by solving (13) (equiv. (11) or (12))
        **repeat**
            Update $G|H, \Theta$ using (14)
            Update $H|G, \Theta$ using (15)
        **until** converged
        **repeat**
            Update $R|C, G, H, \lambda_r$ by optimizing (16)
            Update $C|R, G, H, \lambda_c$ by optimizing (16)
        **until** converged
        Update $\sigma^2|M, G, H$ using (17)
    **until** converged
    **Return** $M, G, H, \Theta$

We apply an alternating optimization approach, alternating between solving for $R^{-1}$ and $C^{-1}$. Each of these sub-optimization problems can be solved using *glasso* (Friedman et al., 2008)

**Noise variance update:** We also may also update the output noise variance. Ignoring terms independent of the noise variance, the optimization is given by minimizing (with respect to $\sigma^2$):

$$ND\log\sigma^2 + \frac{1}{\sigma^2}\left[\|\|Y - XW\|\|_2^2 + \text{tr}(X^\top XG)\,\text{tr}(H)\right]$$

This can be solved in closed form. The solution is given by:

$$\sigma^2 = \frac{1}{ND}\left[\|\|Y - XW\|\|_2^2 + \text{tr}(X^\top XG)\,\text{tr}(H)\right] \quad (17)$$

### 3.3 ALGORITHM

Our goal is to minimize the cost function (4). We solve this by alternating between constrained inference and parameter estimation. Constrained inference involves estimation of the postdata distribution $q(W|Y)$ subject to rank (nuclear norm) and Kronecker covariance constraints, and constrained parameter estimation involves the estimation of updated parameters $\Theta$. The proposed algorithm is summarized in Algorithm 1.

## 4 RELATED WORK

Examples of constrained Bayesian inference in the literature include maximum entropy discrimination (Jaakkola et al., 1999) and posterior regularization Ganchev et al. (2010). Ganchev et al. (2010) applied constrained Bayesian inference techniques to statistical word alignment, multiview learning, dependency parsing and part of speech induction. More recently, researchers have applied constrained Bayesian inference for combining complicated nonparametric topic models with support vector machine inspired large margin constraints for document classification (Zhu et al., 2009), multitask classification (Zhu et al., 2011) and link prediction (Zhu, 2012).

Constrained Bayesian inference is closely related to techniques for approximate variational Bayesian inference (Bishop, 2006), used to approximate intractable Bayesian posterior densities. The approximation typically takes the form of factorization assumptions between subsets of the latent variables. The result is often much easier to solve. Although approximate variational inference also requires solving a constrained version of (3), the motivations and results are quite different than in constrained Bayesian inference methods. In particular, the estimated constrained Bayes distributions may not factorize over subsets of the latent variables.

Partial Least squares (PLS) (Abdi, 2010) is a popular approach for low rank multiple regression. PLS estimates low rank factors that best matches the cross correlation between the features and the response and is known to be especially effective when the feature matrix has co-linear rows and when the features are very high dimensional. Argyriou et al. (2007) and Yuan et al. (2007) proposed models for multitask learning using a regularizer that penalizes the nuclear norm of the weight matrix. This constraint often results in a weight matrix of low rank. Rai & Daumé III (2010) proposed a nonparametric Bayesian model for multitask learning using the direct low rank factor representation. The proposed approach is able to estimate the number of factors using the Indian buffet process prior.

(Zhang & Schneider, 2010; Allen & Tibshirani, 2012) studied covariance estimation for the matrix-variate Gaussian distribution subject to $l_1$ constraints on the precision when the observed data was generated directly from a matrix-variate Gaussian distribution. Stegle et al. (2011) extended the work to the case where the matrix-variate Gaussian distribution is used a the prior, coupled with additive noise. They showed that capturing the additive noise structure can make a significant impact on the quality of the recovered precision matrix. They noted the in difficulty of inference in the model and proposed a heuristic using only the posterior mean and ignoring the posterior covariance. Following our development, the heuristic inference approach of Stegle et al. (2011) can now be explained as a constrained Bayesian inference subject to the constraint the the posterior covariance vanishes. We compare the performance of this heuristic with the Kronecker constrained inference approach on simulated and real data experiments, showing the utility of the richer posterior covariance structure.

## 5 EXPERIMENTS

We present experimental results comparing the proposed rank constrained variational approach to other matrix-variate learning models in the literature. We compared the models in terms of regression accuracy and in terms of structure recovery for the underlying precision matrices. The compared models are as follows:

- Graphical Lasso (GLasso) (Friedman et al., 2008) estimates a sparse precision matrix to match the sample covariance of the response matrix. GLasso was used a the baseline for inter-task structure recovery.

- Multiple regularized ridge regression (Ridge) was used a the baseline for the regression accuracy. Ridge does not estimate the precision matrix.

- Partial least squares (PLS) (Abdi, 2010) estimates low rank factors that best matches the cross correlation between the features and the response. The resulting weight vector can be used for prediction. PLS does not estimate the precision matrix.

- Nuclear norm regularized linear regression (Nuc. Norm) (Yuan et al., 2007) estimates the regression matrix that best predicts the target response subject to a nuclear norm constraint. The resulting weight matrix is often of low rank. Nuc. Norm does not estimate the precision matrix.

- We implemented the matrix variate regression and sparse precision matrix estimation procedure of (Stegle et al., 2011) (MVG). Our approach fixed the feature matrix instead of estimating it from data. As noted in Section 4, Stegle et al. (2011) used a heuristic procedure with a degenerate posterior covariance for the model inference. There is no low rank constraint applied to the model.

- We implemented a *corrected* matrix variate regression and sparse precision matrix estimation procedure using the Kronecker product posterior covariance constraint proposed in Section 3.1 (MVG$_{corr.}$). There is no low rank constraint applied to the model.

- We implemented the proposed nuclear norm constrained matrix variate regression and sparse precision matrix estimation using the constrained Bayesian inference approach regression (MVG$_{rank}$). This combines the nuclear norm constrained inference for the low rank weight matrix with $l_1$ constrained precision matrix learning.

We optimized the proposed MVG$_{rank}$ model by using the approach outlined in Algorithm 1. A similar procedure without the nuclear norm constraint was used to optimize the MVG$_{corr.}$ and MVG models. Nuc. Norm was optimized using a special case of MVG$_{rank}$ without any of the covariance matrices. GLasso, Ridge and PLS were optimized using implementations from the *scikit-learn* python package (Pedregosa et al., 2011).

## 5.1 SIMULATED DATA

Constrained inference is most useful when data is scarce. We are most interested in high dimensional multiple regression where there are more dimensions than samples. In such scenarios, the model constraints can be critical for effective regression and parameter estimation. We performed experiments using simulated data that matches the characteristics of functional neuroimaging data. We fixed the row precision matrix and tested the models ability of estimate the structure of the column precision matrix and the predictive accuracy of the model.

We generated a random row precision matrix by generating using the approach outlined in Example 1 of Li & Toh (2010). We first generated a sparse matrix $\mathbf{U}$ with non zero entries equal to $+1$ or $-1$, then set $\mathbf{C}^{-1} = \mathbf{U}\mathbf{U}^\top$. Finally, we added a diagonal term to ensure $\mathbf{C}^{-1}$ is positive definite. The resulting column precision matrix had a sparsity of 20%. The column precision was generated as the normalized Laplacian matrix (Smola & Kondor, 2003) of a chain graph with a adjacency matrix set as $\mathbf{A}_{i,j} = 1$ if $j = \{i, i+1, i-1\}$ and zero otherwise.

We generated a low rank weight matrix using the factor model as $\mathbf{W} = \mathbf{A}\mathbf{B}^\top$. The columns of $\mathbf{A}$ were generated from the zero mean multivariate Gaussian distribution $\mathcal{N}(\mathbf{0}, \mathbf{R})$ and the columns of $\mathbf{B}$ were generated from the zero mean multivariate Gaussian distribution $\mathcal{N}(\mathbf{0}, \mathbf{C})$. We also generated random high dimensional feature matrices $\mathbf{X} \in R^{N \times D}$ with $x_{i,j} \sim \mathcal{N}(0, 1)$. Finally the response matrix was generated as $\mathbf{Y} = \mathbf{X}\mathbf{W} + \mathbf{N}$ where $\mathbf{N}$ represents independent additive noise with each entry $n_{i,j} \sim \mathcal{N}(0, \sigma^2)$. We selected $\sigma^2$ to maintain a signal to noise ratio of 10.

Our domain of interest is characterized by high dimensional feature variables and few samples. Hence we set the dimensions as $N = 50$, $D = 200$ and $K = 10$. we performed experiments in the low rank regime (rank = 2) and the full rank regime (rank = 10). Experiments were performed using training, validation and test sets with the same number of samples. All experiments were repeated 10 times. The validation set was used for parameter selection. The regularization parameter for all the models except for PLS were selected from the set $\{10^{-3}, 10^{-2}, \ldots 10^3\}$. PLS is not regularized but requires selection of the number of factors. These were chosen from the set $\{2, 4, \ldots 10\}$.

The regression accuracy was measured using the coefficient of determination on the test set. The $R^2$ metric given by $1 - \sum(\hat{y} - y)^2 / \sum(y - \mu)^2$ where $y$ is the target response with sample mean $\mu$ and $\hat{y}$ is the predicted response. $R^2$ measures the gain in predictive accuracy compared to a mean model and has a maximum value of 1. The structure recovery was measured using the area under the *roc* curve (AUC) (Cortes & Mohri, 2004) using the structure of the true precision matrix as the binary target, and the values in the recovered precision matrix as scores. AUC measures the quality of the ranking recovered by by the estimated precision matrix. We also present inference only results with the proposed models using the known precision matrix. The results from the simulated data experiments are shown in Table 1 and Table 2. In both tables, we note that (*) represents inference only results using the true precision matrix.

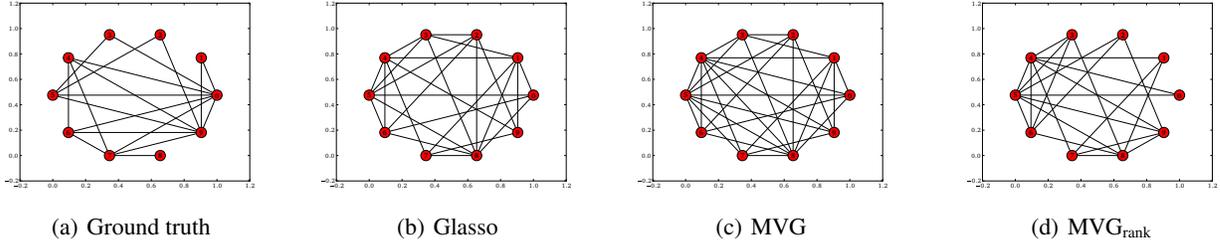

(a) Ground truth  (b) Glasso  (c) MVG  (d) MVG$_{\text{rank}}$

Figure 2: Ground Truth and Recovered Precision Structure with Rank 2 Simulated Data.

Table 1: Average (std.) Accuracy ($R^2$) and Structure Recovery (AUC) for Rank 2 Simulated Data.

| MODEL | $R^2$ | AUC |
|---|---|---|
| GLasso | – | 0.610 (0.071) |
| Ridge | 0.219 (0.029) | – |
| PLS | 0.214 (0.033) | – |
| Nuc. Norm | 0.220 (0.035) | – |
| MVG* | 0.215 (0.033) | – |
| MVG$_{\text{corr.}}$* | 0.271 (0.038) | – |
| MVG$_{\text{rank}}$* | 0.296 (0.038) | – |
| MVG | 0.220 (0.035) | 0.646 (0.048) |
| MVG$_{\text{corr.}}$ | 0.221 (0.035) | 0.648 (0.069) |
| MVG$_{\text{rank}}$ | 0.299 (0.038) | 0.665 (0.064) |

Table 2: Average (std.) Accuracy ($R^2$) and Structure Recovery (AUC) for Rank 10 Simulated Data.

| MODEL | $R^2$ | AUC |
|---|---|---|
| GLasso | – | 0.708 (0.071) |
| Ridge | 0.180 (0.036) | – |
| PLS | 0.169 (0.037) | – |
| Nuc. Norm | 0.168 (0.037) | – |
| MVG* | 0.179 (0.042) | – |
| MVG$_{\text{corr.}}$* | 0.246 (0.035) | – |
| MVG$_{\text{rank}}$* | 0.246 (0.035) | – |
| MVG | 0.172 (0.037) | 0.702 (0.068) |
| MVG$_{\text{corr.}}$ | 0.172 (0.038) | 0.721 (0.071) |
| MVG$_{\text{rank}}$ | 0.245 (0.033) | 0.700 (0.052) |

**Regression:** We found that that accounting for the prior correlation structure had a significant effect on the quality of the recovered regression. Hence, although the Nuc. Norm model performed better than Ridge, models that combined regression with structure recovery outperformed models using regression only. The corrected MVG$_{\text{corr.}}$ outperformed the MVG in regression suggesting the importance of capturing the posterior covariance for regression and parameter estimation performance. We note that even when the rank is full, the underlying weight matrix is given by the product of the factors is not Gaussian distributed. This may account for the observation that the Gaussian based models perform worse for the full rank data. Another reason may be the significant increase in the effective dimensionality of the weight matrix parameter to be estimated using the same amount of data. MVG$_{\text{rank}}$ is able to compensate for this mismatch.

**Structure recovery:** Overall, all models improved accuracy of recovery for the precision structure as the rank was increased. At the low rank, the MVG$_{\text{rank}}$ model was the most effective for structure recovery, but the MVG$_{\text{corr.}}$ model was the most effective at high rank. We also found that correcting the inference procedure improved the structure recovery performance by comparing MVG$_{\text{corr.}}$ to MVG. We counted the number of times each edge was selected over the random repetitions. We present the recovered graphs for rank 2 simulated data showing links selected in at least 70% of the repetitions with weight greater than $10^{-6}$ in Fig. 2. MVG selects many more edges than the MVG$_{\text{corr.}}$ method in this experiment.

### 5.2 FUNCTIONAL NEUROIMAGING DATA

Functional magnetic resonance imaging (fMRI) is an important tool for non-invasive study of brain activity. Most fMRI studies involve measurements of blood oxygenation (which are sensitive to the amount of local neuronal activity) while the participant is presented with a stimulus or cognitive task. Neuroimaging signals are then analyzed to identify which brain regions exhibit a systematic response to the stimulation, and thus to infer the functional properties of those brain regions. Functional neuroimaging datasets typically consist of a relatively small number of correlated high dimensional brain images. Hence, capturing the inherent structural properties of the imaging data is critical for robust inference.

We completed experiments using brain image data from an extended set of the openfMRI database[2]. The data was preprocessed using a general linear model with FMRIB Software Library (FSL) to compute contrast images for each subject resulting in $N = 479$ contrast images for $K = 26$ contrasts. The target contrasts were encoded into a response matrix using the 1-of-$k$ representation, where $y_{n,k} = 1$ if image $n$ corresponds to task $k$ and is zero otherwise. after masking, we are left with $D = 174264$ dimensions. Each dimension in the brain image corresponds to a spatial location in the brain. We used the normalized Laplacian of the 3-dimensional spatial graph of the brain image voxels to define the row precision matrix. This corresponds to the observation that nearby voxels tend to have similar functional activation. Our approach is motivated by the observation that functional neuroimages are highly correlated for different tasks (Poldrack, 2011). We seek to extract this correlation structure as encoded in the prior precision matrix. In addition, the high dimensionality and the similarity between different tasks suggests that the optimal weight matrix may be of low rank.

We divided the training data into five sets using a stratified cross validation to ensure that each training set contains a similar relative number of images corresponding to each task. In addition to the proposed models, we present experimental results using the support vector machine classifier (SVM) using implementations from the *scikit-learn* python package (Pedregosa et al., 2011). We also note that the ridge regression is exactly equivalent to the least square support vector machine (LS-SVM) (Ye & Xiong, 2007) with a linear kernel. For all models (except for PLS ), we selected the regularization parameter from the set $\{10^{-3}, 10^{-2}, \ldots 10^3\}$. The number of factors in PLS was selected from the set $\{2, 4, 6 \ldots 26\}$. The results are provided in Table 3.

Table 3: Average (std.) Classification Accuracy for fMRI Data

| MODEL | ACCURACY |
| --- | --- |
| SVM | 0.463 (0.052) |
| PLS | 0.422 (0.030) |
| LS-SVM | 0.472 (0.040) |
| Nuc. Norm | 0.234 (0.022) |
| MVG | 0.463 (0.052) |
| MVG$_{\text{corr.}}$ | 0.476 (0.050) |
| MVG$_{\text{rank}}$ | **0.512 (0.034)** |

---

[2] https://openfmri.org/, extended data provided courtesy of openfMRI.

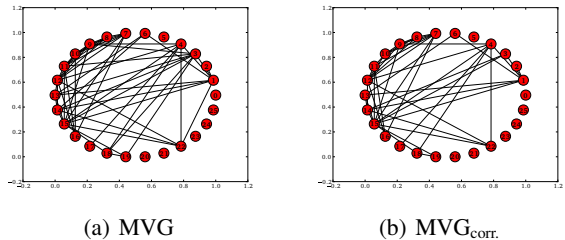

(a) MVG  (b) MVG$_{\text{corr.}}$

Figure 3: Recovered Precision Structure for fMRI Data

We note the difficulty of this classification task due to the large number of classes and the high dimensionality of the features. Fig. 3 compares the most significant edges recovered by the precision matrices of MVG and MVG$_{\text{corr.}}$ methods. The figure shows edges with absolute value of the weight greater than the $90^{th}$ percentile. We found that MVG selected more edges than the MVG$_{\text{corr.}}$ method. We are in the process of collaborating with domain experts for further analysis of the task similarities encoded by the the task precision matrices. These results will be included in an extended version of the paper.

## 6 CONCLUSION

We proposed a novel primal approach for Bayesian inference subject to possibly non-convex constraints. We applied the proposed inference approach to rank constrained multitask learning. Our approach was motivated by an application to reverse inference for high dimensional functional neuroimaging data. We developed an algorithm for constrained inference that accounts for the latent structure of the predictive weight matrix and constrained parameter estimation to learn the sparse conditional independence structure between the tasks as encoded by the prior precision matrices. We presented experimental performance results compared to strong baseline models on simulated data and real functional neuroimaging data.

We are interested in extending the proposed approach to constrained inference for nonparametric Bayesian models. In particular, we are interested in rank constrained models for the matrix-variate Gaussian process applied to matrix completion. We are also interested in further theoretical development to understand the trade-offs of constrained Bayesian inference compared to other approaches.

### Acknowledgments

Authors acknowledge support from NSF grant IIS 1016614. We thank Russell A. Poldrack for providing the extended openfMRI dataset and for help with preprocessing. We also thank Sreangsu Acharyya and Suriya Gunasekar for helpful suggestions on early drafts of this manuscript.